\pdfoutput=1
\documentclass[11pt]{article}

\usepackage[]{EMNLP2022}

\usepackage{times}
\usepackage{latexsym}

\usepackage[T1]{fontenc}
\usepackage[utf8]{inputenc}

\usepackage{microtype}

\usepackage{amsfonts}
\usepackage{booktabs}
\usepackage{array,multirow}

\usepackage{float}
\usepackage{soul}
\usepackage{mathtools}
\usepackage{varwidth}
\usepackage{capt-of}% or \usepackage{caption}
\usepackage{subcaption}
\usepackage{makecell}
\usepackage{enumitem}

\usepackage{xcolor,colortbl}

\usepackage{algorithmic}
\usepackage[linesnumbered,ruled]{algorithm2e}

\SetCommentSty{mycommfont}
\SetKwComment{Comment}{$\triangleright$\ }{}

\newcommand{\MD}{{\small\textsf{MSRP}}}
\newcommand{\FMD}{Multi-Summary based Reinforcement learning with Pretraining}

\definecolor{gred}{RGB}{219,68,55}
\definecolor{gblue}{RGB}{66,133,244}
\definecolor{gyellow}{RGB}{244,180,0}
\definecolor{ggreen}{RGB}{15,157,88}
\definecolor{ggrey}{RGB}{115,115,115}

\newcommand{\red}[1]{\textcolor{gred}{\textbf{#1}}} 
\newcommand{\blue}[1]{\textcolor{gblue}{\textbf{#1}}}

\setstcolor{ggrey}

\title{Generating Multiple-Length Summaries via Reinforcement Learning \\ for Unsupervised Sentence Summarization}

\author{Dongmin Hyun$^{\spadesuit}$\thanks{~~Work done during internship at Microsoft Research Asia.} \hspace{3mm} Xiting Wang$^{\heartsuit}$ \hspace{3mm} Chanyoung Park$^{\clubsuit}$ \\ \textbf{Xing Xie}$^{\heartsuit}$ \hspace{3mm} \textbf{Hwanjo Yu}$^{\spadesuit}$\thanks{~~Corresponding author} \vspace{1mm} \\
$^{\spadesuit}$Pohang University of Science and Technology \hspace{1mm} $^{\heartsuit}$Microsoft Research Asia \\
$^{\clubsuit}$Korea Advanced Institute of Science and Technology
\\
\small{\texttt{\{dm.hyun, hwanjoyu\}@postech.ac.kr}} \hspace{1mm} \small{\texttt{cy.park@kaist.ac.kr}}\\
\small{\texttt{\{xitwan, xing.xie\}@microsoft.com}}
}

\begin{document}
\maketitle
\begin{abstract}
Sentence summarization shortens given texts while maintaining core contents of the texts. Unsupervised approaches have been studied to summarize texts without human-written summaries. However, recent unsupervised models are extractive, which remove words from texts and thus they are less flexible than abstractive summarization.   
In this work, we devise an abstractive model based on reinforcement learning without ground-truth summaries. 
We formulate the unsupervised summarization based on the Markov decision process with rewards representing the summary quality. 
To further enhance the summary quality, we develop a multi-summary learning mechanism that generates multiple summaries with varying lengths for a given text, while making the summaries mutually enhance each other. 
Experimental results show that the proposed model substantially outperforms both abstractive and extractive models, yet frequently generating new words not contained in input texts.
\end{abstract}

\section{Introduction}

The goal of sentence summarization is to enhance the readability of texts by reducing their lengths through word dropping, replacement, or paraphrasing. The applications of the task include subtitle generation \cite{luotolahti2015sentence} and email summarization \cite{zajic2008single}. An issue is that it is costly to have human editors write summaries for each text. Hence, it is critical to develop an unsupervised model that does not require any human-written summaries.

Early models focus on abstractive summarization that \textit{generates} words from a vocabulary set rather than extractive summarization, which merely \textit{selects} words from texts.
Specifically, abstractive models have adopted autoencoder networks to summarize texts in an unsupervised manner \cite{wang2018learning,fevry2018unsupervised,baziotis2019seq}. 
In contrast, extractive models summarize texts by finding word combinations from texts, aiming at maximizing predefined scores (e.g., fluency of summaries) \cite{west2019bottlesum}. Despite their limited functionality, i.e., word selection, recent extractive models outperformed the abstractive models \cite{schumann2020discrete,liu-etal-2022-learning}. 

Despite the success of the extractive models, we argue that they have an inherent downside. 
The extractive models only select words from texts, and thus they cannot generate new words that can be effective for sentence summarization. 
For example, extractive models are unable to generate acronyms (e.g., PM) for words (e.g., Prime Minister) if the acronyms do not appear in texts.
In contrast, abstractive models can resolve the limitation of extractive models. However, the summary quality of existing abstractive models is sometimes worse than a simple baseline, which simply truncates input texts from the beginning \cite{schumann2020discrete}. This implies that existing abstractive models fall short of reducing text lengths while maintaining the summary quality.
The aforesaid limitations of existing models motivate us to devise an abstractive model that produces high-quality summaries with generating new words not contained in input texts.

This work employs reinforcement learning (RL) for unsupervised abstractive summarization\footnote{Public source codes: \href{https://github.com/dmhyun/MSRP}{\color{magenta}{\textit{https://github.com/dmhyun/MSRP}}}}. RL enables a model to learn to summarize using rewards even though they are non-differentiable. 
Our model generates \textit{high-quality} summaries with considering 1) the semantic similarity between the generated summary and its corresponding input text, and 2) fluency of the generated summaries. 
Notably, the semantic similarity is more robust to preserve core contents of input texts than the word-level reconstruction objective \cite{pgj2017unsup}, which is adopted by existing abstractive models. 

Moreover, we argue that the difficulty of summarization depends on the summary lengths (e.g., the shorter the summary, the more difficult it is to summarize). 
In this respect, we develop a \textit{multi-summary learning} mechanism that generates multiple summaries with varying lengths for a given text, while making the summaries mutually enhance each other. 
The main idea is to use a high-quality summary of a certain length, which is \textit{easy} to generate, to enhance the quality of a low-quality summary of another length, which is \textit{difficult} to generate, rather than independently generating summaries in each length. 
Specifically, we design the mechanism to make low-quality summaries semantically similar to high-quality ones.

We also devise a pretraining task to obtain well-initialized model parameters for the RL training.
We first augment input texts by applying word-level perturbations and inserting length prompts, which indicate the lengths of the original texts.
Then, we train the model to reconstruct the original text from the augmented one, which makes the model learn to summarize and control the output length. 
By pretraining the model in this manner, our goal is to equip the model with essential abilities for summarization, which results in an improved summary quality after the RL training with the pretrained model.
We dub our model \FMD{} (\MD).

Experiments show that \MD{} outperforms the abstractive and extractive baseline models in both automatic and human evaluations. 
We also analyze summaries generated by \MD{} to illuminate its benefits compared to the recent extractive models. 

\section{Related Work}
\subsection{Unsupervised Sentence Summarization} 
Supervised models depend on human-written summaries, which involve costly and time-consuming data creation \cite{rush-etal-2015-neural,he2020ctrlsum,song2021new}. In contrast, unsupervised models learn to summarize texts without any human-written summaries. 
Abstractive models mainly adopt autoencoders to build a summarization model.
\citet{fevry2018unsupervised} adopt a denoising autoencoder to summarize texts by treating texts as noised data and summaries as clean data. 
\citet{wang2018learning,baziotis2019seq} design autoencoders that generate word sequences as interim outputs of the autoencoders and use the word sequences as summaries. 
\citet{zhou2019simple} devise a model that selects the best next word based on a fluency score to generate summaries.
In contrast, an extractive model \cite{west2019bottlesum} iteratively deletes words from texts to generate summaries while measuring the fluency of each intermediate summary. \citet{schumann2020discrete} select the best word combination that maximizes predefined scores based on a hill-climbing search algorithm, and it surpassed the abstractive models. However, the search requires exhaustive computation. In response, \citet{liu-etal-2022-learning} train an extractive model with summaries generated by \citet{schumann2020discrete} so that it can quickly generate summaries without the exhaustive search.
Compared to extractive models, this work aims to design an abstractive model to enjoy its flexible operation, i.e., generating words not contained in texts.

\subsection{Reinforced Summarization Models}
RL has been used as a technique to solve summarization tasks. 
With referential summaries, 
\citet{paulus2018deep,bian2019controllable} relieve the exposure bias of teacher forcing-based supervision.
Without referential summaries, \citet{bohm-etal-2019-better,stiennon2020learning} devise RL-based models that maximize a reward representing the summary quality, which is annotated by human experts. 
\citet{wang2018learning} address the unsupervised sentence summarization where only input texts are available, which is our target scenario. 
They utilize RL to train an autoencoder with a word-level reconstruction loss to preserve contents of texts in summaries. 
In this work, we formulate a RL framework to achieve three aspects: 1) semantic similarity between input texts and summaries instead of word-level similarity, 2) controllability on summary length, and 3) model-agnostic RL framework. 

\subsection{Pretraining Task for Summarization}
Pretraining tasks are crucial to obtain high accuracy on NLP tasks \cite{devlin2019bert, lewis-etal-2020-bart}. 
Recent research invents pretraining tasks for long-document summarization \cite{zhang2020pegasus,zhu2021leveraging}. 
However, the approaches are not applicable to sentence summarization due to the absence of multiple sentences, and do not consider controlling the summary length. We thus propose an effective pretraining task to make models learn to summarize and control the summary length.

\section{Method}

\subsection{Problem Formulation}
The goal of sentence summarization is to shorten a text (i.e., a long sentence) $\textbf{t}=[w_1, w_2, \cdots, w_{|\textbf{t}|}]$ into a short summary $\textbf{y}=[y_1, y_2, \cdots, y_{|\textbf{y}|}]$ where $w$, $y$ are words and $|\textbf{y}| < |\textbf{t}|$.
It is important to note that the text-summary pairs are not available for training models. In other words, we focus on the unsupervised sentence summarization.

\subsection{Reinforcement Learning Framework} 

\begin{figure}[t]
	\centering
	\includegraphics[width=.65\linewidth]{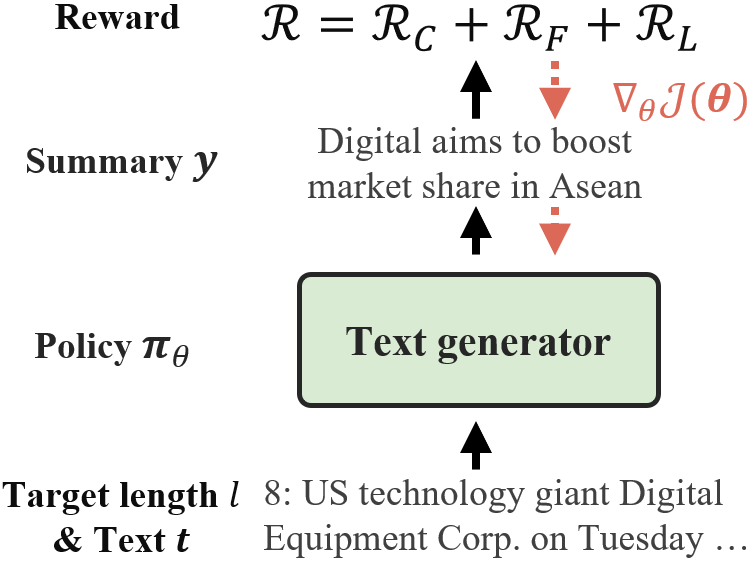} %.7
	\caption{Reinforcement learning with a length prompt. 
	}
	\label{fig:rl}
\end{figure}

Due to the absence of ground-truth summaries, we train a text generator based on the quality of generated summaries. However, the summary generation requires the word-sampling process, which is non-differentiable. We thus consider RL to address the non-differentiability (Figure \ref{fig:rl}), and describe the proposed Markov decision process as follows.

\textbf{States} describe the possible combinations of input texts $\textbf{t}$ and generated summaries $\textbf{y}_t = [y_1, y_2, \cdots, y_t] $ at time $t$. 
State at time $t$ can be formulated as $s_t=[\textbf{t}, \textbf{y}_t]$. 
\textbf{Actions} are the candidate next words from a vocabulary set $\mathcal{V}$ at given states.
A policy $\boldsymbol{\pi}_\theta$ selects an action $a_t \in \mathcal{V}$ as a next word $y_{t+1}$ based on a given state $s_t$, resulting in next summary $\textbf{y}_{t+1}$.
\textbf{Transition function} determines next states based on a state $s_t$ and action $a_t$, i.e., $s_{t+1} = \mathcal{T}(s_t, a_t)=[\textbf{t}, \textbf{y}_{t+1}]$.

\noindent
\textbf{Reward} $\mathcal{R}(s_t, a_t)$ represents the summary quality when a target summary length $l$ is given. 
We obtain the reward of the generated summaries such that: 
\begin{align*}
  \small
  \mathcal{R}(s_t, a_t)=\begin{cases}
    \mathcal{R}(\textbf{y}, \textbf{t}, l) & \text{if $a_t=\mathtt{[EOS]}$ $\vee$ $t=M_g$},\\
    0 & \text{otherwise,} 
  \end{cases}
\end{align*}
where $\textbf{y}$ denotes the generated summary (i.e., $\textbf{y} = \textbf{y}_t$ for simplicity), $\mathtt{[EOS]}$ is the end-of-sentence token, $M_g$ is the maximum length of generated summaries. 
We design pertinent aspects of summaries:
\begin{align*}
    \mathcal{R}(\textbf{y}, \textbf{t}, l) =  \mathcal{R}_{C}(\textbf{y}, \textbf{t}) + \mathcal{R}_{F}(\textbf{y}) + \mathcal{R}_{L}(|\textbf{y}|, l). 
\end{align*}

\vspace{-1ex}

\begin{itemize}[leftmargin=*]
    \item \textbf{Content preservation} A requirement for high-quality summaries is to preserve the gist of the input texts. We consider the semantic similarity between summaries and the corresponding texts:
    \begin{equation}
        \mathcal{R}_{C}(\textbf{y}, \textbf{t}) = \mathtt{sim}(f(\textbf{y}), f(\textbf{t}))
        \label{eq:reward_cp}
    \end{equation}
    where $\mathcal{R}_C \in [0,1]$, $\mathtt{sim}$ is a similarity function, and $f$ is a function to embed texts (i.e., $\textbf{y}$ and $\textbf{t}$) such as BERT.\footnote{The specific model is described in Section \ref{sec:imp_detail}.} We use cosine similarity with normalization, i.e., $\mathtt{sim}(\cdot, \cdot) = (\mathtt{cos}(\cdot, \cdot)+1)/2$. 
    The semantic similarity enables the model to robustly capture the meaning of texts despite different words in two texts, e.g., \textit{Who's the winner?} and \textit{Who won the game?}.
    
    \item \textbf{Fluency} Another requisite for summaries is fluency, representing how generated summaries are grammatically and semantically natural. 
    We use the perplexity as fluency:
    \begin{equation}
        \mathtt{PPL}(\textbf{y}) = \exp \big\{ -\frac{1}{|\textbf{y}|} \sum_t^{|\textbf{y}|} \log p_\psi(\text{y}_t| \textbf{y}_{t-1}) \big\}
        \label{eq:reward_fl}
    \end{equation}
    where $\mathtt{PPL}$ is the perplexity from a language model with its parameters $\psi$, and $\textbf{y}_{t-1}$ is the generated summary before time $t$. Low $\mathtt{PPL}$ indicates high fluency.
    We define the fluency reward:
    \begin{align}
        \mathcal{R}_{F}(\textbf{y}) = \exp(-\mathtt{PPL}(\textbf{y}) / \sigma_F)
        \label{eq:R_F}
    \end{align}
    where $\mathcal{R}_F \in (0,1]$ and $\sigma_F \in \mathbb{R}_+$ is a tunable scaling factor to control the steepness of $\mathcal{R}_{F}$.\footnote{The shape of the reward is provided in Appendix \ref{apnd:scale}.} 
    
    \item \textbf{Summary length} We design our model to summarize texts in a desired length. We first insert the desired length $l$ (e.g., 8 words) at the beginning of an input text (e.g., `8:' Figure \ref{fig:rl}), and then optimize the following reward:
    \begin{align*}
        \mathcal{R}_{L}(|\textbf{y}|, l) = \exp(-||\textbf{y}| - l|/\sigma_L)
    \end{align*}
    where $\mathcal{R}_{L}\in (0, 1]$ and $\sigma_L \in \mathbb{R}_+$ is a tunable scaling factor. 
    After training, we can control the summary length by changing the desired length.     
\end{itemize}

\subsubsection{Policy Gradient}
Policy gradient directly updates the policy parameters $\theta$ to minimize an objective function $\mathcal{J}$:
\begin{align*}
    \small
    \mathcal{J}(\theta)=-\sum_{\textbf{t}\in T} \mathbb{E}_{\textbf{y} \small{\sim} \boldsymbol{\pi}_\theta(\cdot|l, \textbf{t})} \mathcal{R}(\textbf{y}, \textbf{t}, l).
\end{align*}
where $T$ is a set of input texts.
Therefore, RL updates the policy parameters $\theta$ to maximize the expected rewards (i.e., $\mathcal{R}_L, \mathcal{R}_C, \text{and } \mathcal{R}_F$).
We adopt a self-critical policy gradient \cite{rennie2017self} to stabilize the training by reducing the variance. The gradient of the policy can be written as:
\begin{align*}
    \scriptsize
    \bigtriangledown_\theta \mathcal{J}(\theta) \approx - \sum_{\textbf{t}\in T}\big (\mathcal{R}(\textbf{y}, \textbf{t}, l) - \mathcal{R}(\bar{\textbf{y}}, \textbf{t}, l) \big ) \bigtriangledown_\theta \sum_t \log \boldsymbol{\pi}_\theta(y_{t+1}|s_t) 
\end{align*}
where $\bar{\textbf{y}}$ is a baseline summary, whose words are greedily selected, i.e., $y_{t+1} = \small \mathrm{argmax}\, \boldsymbol{\pi}_\theta(\text{y}_{t+1}|s_t)$, instead of sampling words, i.e., $y_{t+1} \small{\sim} \boldsymbol{\pi}_\theta(\text{y}_{t+1}|s_t)$. 
The gradient has a direction to maximize the likelihood if  $\mathcal{R}(\textbf{y}, \textbf{t}, l) > \mathcal{R}(\bar{\textbf{y}}, \textbf{t}, l)$. 

\begin{figure}[t]
	\centering
	\includegraphics[width=.65\linewidth]{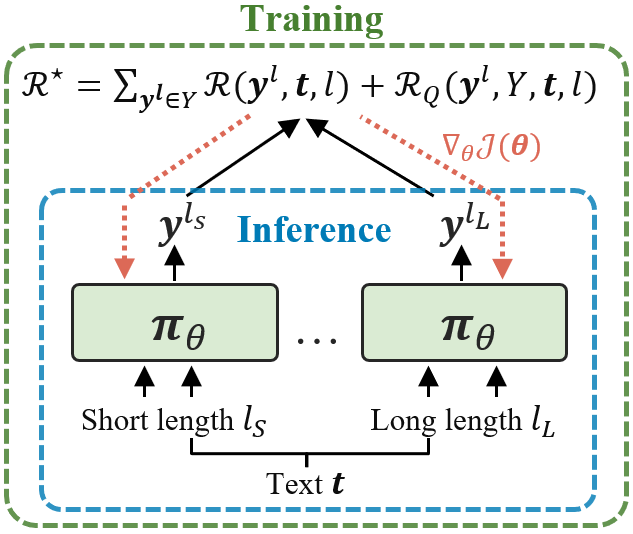} 
	\caption{Multi-summary learning mechanism}
	\label{fig:msl}
\end{figure}

\subsection{Multi-Summary Learning Mechanism}
\label{sec:msl}
We further improve the summary quality by making multiple summaries with varying lengths mutually enhance each other (Figure \ref{fig:msl}).
The main idea is to use a high-quality summary of a certain length, which is \textit{easy} to generate, to enhance the quality of a low-quality summary of another length, which is \textit{difficult} to generate.
We first generate multiple summaries in different lengths for each text: 
\begin{align*}
    Y = \{\textbf{y}^l\}_{l \in L} \; \text{   where   } \; \textbf{y}^l \small{\sim} \boldsymbol{\pi}_\theta(\cdot|l, \textbf{t}),
\end{align*}
where $Y$ is the set of summaries generated for each length $l \in L$, $L$ is a set of lengths, and $\textbf{y}^l$ is a summary generated for the length $l$. For brevity, we denote a target summary as $\textbf{y}$, while the other summaries as $\textbf{y}'$ henceforth.

We then design the mechanism to make a summary semantically similar to the other summaries based on mutual relationship:  \par\nobreak \vspace{-1ex}
{\small
\begin{align*}
   \mathcal{R}_Q(\textbf{y}, Y, \textbf{t}, l) = \lambda \sum_{\textbf{y}' \in Y \setminus \textbf{y}} u(\textbf{y}, \textbf{y}', \textbf{t}, l) \cdot \mathcal{R}_C(\textbf{y}, \textbf{y}')
\end{align*}
}
where $\mathcal{R}_Q \in [0,1]$,  $\lambda \in [0,1]$ is a weight coefficient for this reward,  $u\in [0,1]$ is a function that measures the usefulness of a summary $\textbf{y}'$ to a target summary $\textbf{y}$ generated for a target length $l$ given text $t$. 
Hence, the model makes a target summary $\textbf{y}$ to be semantically similar to another summary $\textbf{y}'$ based on its usefulness, i.e., refer to a summary $\textbf{y}'$ if it is useful to a target summary $\textbf{y}$.

We design the usefulness function $u$ by considering the summary quality and length: 1) given the input text $\textbf{t}$, a target summary $\textbf{y}$ should refer to another summary $\textbf{y}'$ with different length only if the quality of $\textbf{y}'$ is higher than that of $\textbf{y}$, i.e., $q(\textbf{y}', $\textbf{t}$) > q(\textbf{y}, $\textbf{t}$)$, where $q \in [0,1]$ is a function that measures the summary quality.
2) A summary generated for a length $l$ should refer to another summary with similar length to the target length $l$, i.e., the more similar the length, the higher the relevance.
We define the usefulness function $u$: \par\nobreak \vspace{-1ex}
{\small
\begin{align*}
    u(\textbf{y}, \textbf{y}', \textbf{t}, l) =  [q(\textbf{y}', \textbf{t}) - q(\textbf{y}, \textbf{t})]_+^\alpha  \cdot \mathcal{R}_L(|\textbf{y}'|, l)
\end{align*}
}
where $\alpha \in \mathbb{R}$ is a scaling factor, and $[\, \cdot \,]_+$ represents $\max(\cdot, 0)$.
The first term produces a positive score if $q(\textbf{y}', $\textbf{t}$) > q(\textbf{y}, $\textbf{t}$)$. Similarly, second term produces a high score if the length of another summary  $\textbf{y}'$ is close to a target length $l$ of a given summary $\textbf{y}$.
We consider the summary quality based on the content preservation and fluency:
\begin{align*}
    \small
    q(\textbf{y}, \textbf{t}) = \mathcal{R}_C(\textbf{y}, \textbf{t}) \cdot \mathcal{R}_F(\textbf{y}).
\end{align*}

Finally, the total reward $\mathcal{R}^\star$ can be written with multiple summaries and the quality reward $\mathcal{R}_Q$: \par\nobreak \vspace{-1ex}
{\small
\begin{align*}    
    \mathcal{R}^\star(Y, \textbf{t}) = \sum_{\textbf{y}^l\in Y} \mathcal{R}(\textbf{y}^l, \textbf{t}, l) + \mathcal{R}_Q(\textbf{y}^l, Y, \textbf{t}, l).    
\end{align*}

}
This mechanism makes summaries mutually enhance each other during training time, but generates summaries independently in inference time. Thus, the complexity of the inference does not increase. 

\begin{table}[]
    \footnotesize
    \centering
    \scalebox{0.8}{
        \begin{tabular}{p{1\columnwidth}}
        \toprule
            \textbf{\text{Original text}}  \\
            three researchers on monday won the nobel medicine prize for discovering how nitric oxide acts as a signal molecule
              \\
            \midrule            
            \textbf{\textit{1. Shuffle}}  \\
            \red{nitric} \red{three} researchers on monday won the nobel prize for discovering how signal \red{medicine} oxide acts as a  molecule  \\
            
            \vspace{-0.5ex}
            \textbf{\textit{2. Drop}} \\
            \red{\st{nitric}} three researchers on \red{\st{monday}} won the nobel \red{\st{prize}} for discovering \red{\st{how}} signal medicine oxide acts as a molecule \\
            
            \vspace{-0.5ex}
            \textbf{\textit{3. Add \& Prompt}}  \\
            $\overbrace{\,\, \red{20: }}^{\blue{Prompt}}$ three \red{crashing} \red{flight} researchers \red{town} on \red{103} won the \red{down} nobel medicine \red{on} prize for \red{tiny} how \red{this} nitric oxide as a signal molecule \\ 
         \bottomrule
        \end{tabular}
    }
    \caption{Example of text perturbation with a length prompt. Changes in each step are marked in red.}
    \label{tab:pretrain}
\end{table}

\subsection{Pretraining Task}
We also devise \textit{prompt-based text reconstruction} task (Table \ref{tab:pretrain}), and its main goal is to make our model learn to control the output length.
We first apply perturbations to texts: shuffling, dropping, and adding words.
We then insert the length of the original text at the beginning of the perturbed text, called \textit{prompt}, e.g., `20:' in Table \ref{tab:pretrain}.
Thus, by inserting the prompt to perturbed texts, the model can be explicitly informed about the target length for the original text.
We train our model to reconstruct the original text from the perturbed text, which makes the model learn to control the output length and reorder, add, and remove words.
After pretraining, we perform the RL training with the pretrained model.
We provide the details in Appendix \ref{apnd:ptr}.

\section{Experiments}
\subsection{Experimental Settings}
\subsubsection{Datasets}
We evaluate \MD{} on benchmark datasets for sentence summarization. The Gigaword dataset contains a news headline per news article. The number of training and evaluation data are 3,803,957 and 1,951, respectively. We only use news articles to train \MD{} so that our model does not draw on any article-headline pairs. 
We select 500 validation data only for tuning the hyperparameters as done in prior work \cite{schumann2020discrete,liu-etal-2022-learning}.
We also use DUC2004 dataset, designed only for evaluation, consisting of four headlines per news article, and it contains 500 news articles.

\begin{table*}[t]
\footnotesize
\centering
\renewcommand{\arraystretch}{1.0}
\scalebox{0.97}{
\begin{tabular}{ccc|cccc|cc|c}
\toprule
       Group  & Type                & Model                   & RF-1                           & RF-2                          & RF-L   & $\Delta$R & Fidelity & Fluency & Len.                                                           \\ \midrule 
           
        & Ext.                   & Lead (8 words)                            & 21.40                         & 7.43                         & 20.04              & 18.48    & 0.856     & 0.723      & 7.9                                                  \\
                
        & Ext.                    & \citet{schumann2020discrete}                  & 26.01                         & 9.64                         & 23.94           & 7.76     &  0.836    &  \textbf{0.914}            & 7.9                                                  \\
        & Ext.                    & \citet{su2021non}                   & 26.88                         & 9.37                         & 24.54           &  6.56   &   0.817   &  0.883            & 7.7                                                 \\          
        & Ext.                    & \citet{liu-etal-2022-learning}                   & 27.94                         & 9.24                         & 25.51           &   4.66     &  0.857    & 0.760         & 7.8                                                 \\
        & Ext.                    & \citet{liu-etal-2022-learning}$^\dagger$                   & 26.94                         & 9.97                         & 24.93          &  5.51      &  0.847    & 0.878     & 7.9                                                 \\

       \multirowcell{-6}{A\\(desired \\ length 8)} & Abst. & \cellcolor[HTML]{EFEFEF}$\text{\MD}$ \hspace{.01ex}  & \cellcolor[HTML]{EFEFEF}\textbf{29.09} & \cellcolor[HTML]{EFEFEF}\textbf{11.46} & \cellcolor[HTML]{EFEFEF}\textbf{26.80} & \cellcolor[HTML]{EFEFEF}0.0 & \cellcolor[HTML]{EFEFEF}\textbf{0.875} &\cellcolor[HTML]{EFEFEF}0.899    & \cellcolor[HTML]{EFEFEF}7.9  \\  
       
       & Abst.                    & \MD{} w/o RL                    &    23.54                      &    8.36                      &  21.93          &   13.52     &  0.855    &  0.766        &   7.8                                               \\   
       
       \midrule

         & Ext.               & Lead (10 words)                           & 23.04                         & 7.96                         & 21.30             &  16.62        & 0.884     &  0.729      & 9.8                                                   \\
       & Abst.                 & \citet{wang2018learning}             & 27.29                         & 10.01                         & 24.59          &   7.03        &   $-$   &     $-$    & 10.8                                                 \\
       & Abst.                 & \citet{zhou2019simple}           & 26.51                         & 10.04                         & 24.45            & 7.92        &  	0.850  &   0.900      & 9.3                                               \\
       
      & Ext.                  & \citet{schumann2020discrete}            & 27.03                         & 10.13                        & 24.61            &  7.15        &  0.856    &  \textbf{0.914}      & 9.8                                                \\
     & Ext.                    & \citet{su2021non}                   & 27.86                         & 9.88                         & 25.51           &   5.64        &  0.832     &  0.889      & 9.4                                                 \\                  
      & Ext.                  & \citet{liu-etal-2022-learning}              & 28.55                         & 9.97                        & 25.78            &   4.62         &  0.873    &   0.798    & 9.8                                                 \\
      & Ext.                  & \citet{liu-etal-2022-learning}$^\dagger$              & 27.61                         & 10.23                        & 25.04            &    6.04        &  0.865    &   0.848    & 9.8                                                 \\

\multirowcell{-8}{B\\(desired \\ length 10)} & Abst.  &  \cellcolor[HTML]{EFEFEF}$\text{\MD}$ \hspace{.01ex}   & \cellcolor[HTML]{EFEFEF}\textbf{29.94} & \cellcolor[HTML]{EFEFEF}\textbf{11.86} & \cellcolor[HTML]{EFEFEF}\textbf{27.12} &  \cellcolor[HTML]{EFEFEF}0.0 &  \cellcolor[HTML]{EFEFEF}\textbf{0.897}    &   \cellcolor[HTML]{EFEFEF}0.886   & \cellcolor[HTML]{EFEFEF}9.9  \\ 

& Abst.                    & \MD{} w/o RL                    &    25.01                      &  8.86                        &   22.95         &  12.10      & 0.885     &   0.751       & 9.9 \\ 
\midrule

          & Ext.               & Lead (50\% words)                           & 24.97                         & 8.65                         & 22.43             & 8.72  & 0.917 & 0.739         & 14.6                                                 \\
          & Abst.                 & \citet{fevry2018unsupervised}             & 23.16                         & 5.93                         & 20.11          & 15.57  & $-$ & $-$             & 14.8                                                 \\
           & Abst.                 & \citet{baziotis2019seq}           & 25.49                         & 8.27                         & 22.76            & 8.25    & 0.919 &  0.680         & 14.9                                               \\
                                   
           & Ext.                  & \citet{schumann2020discrete}            & 27.05                         & 9.75                        & 23.89            & 4.08  &$-$ &   $-$         & 14.9                                               \\           
           & Ext.                  & \citet{liu-etal-2022-learning}             & 28.53                         & 9.88                        & 25.10            & 1.26  & 0.901 &   0.789         & 14.9                                                \\

\multirowcell{-7}{C\\ (desired length \\ 50\% of the input)} & Abst.  &  \cellcolor[HTML]{EFEFEF}$\text{\MD}$ \hspace{.01ex}   & \cellcolor[HTML]{EFEFEF}\textbf{28.60} & \cellcolor[HTML]{EFEFEF}\textbf{11.00} & \cellcolor[HTML]{EFEFEF}\textbf{25.17} &  \cellcolor[HTML]{EFEFEF}0.0 & \cellcolor[HTML]{EFEFEF}\textbf{0.924} & \cellcolor[HTML]{EFEFEF}\textbf{0.795} &  \cellcolor[HTML]{EFEFEF}14.8\\ 
& Abst.                    & \MD{} w/o RL                    &   26.40                       &  9.37                        &  23.76          &  5.24      & 0.921     &   0.715       & 14.4  \\
\bottomrule
\end{tabular}
}
\caption{Automatic evaluation on Gigaword dataset. $\Delta_R$: the improvement of total ROUGE of \MD{} over each model, Len: averaged length of summaries, $^\dagger$:  \citet{liu-etal-2022-learning} with the same pretrained model used for \MD{}.} 
\label{tab:rouge}
\end{table*}

\begin{table}[]
    \centering
    \resizebox{0.9999\linewidth}{!}{
    \begin{tabular}{cc|cccc|ccc}
         \toprule
         \multicolumn{8}{c}{Group D (desired length 13)} \\ 
         
           Type     & Model  & RR-1 & RR-2 & RR-L & $\Delta$R & FD & FL   \\ \midrule
           Ext.                   & Lead (75 char.)                          & 22.54                         & 6.52                         & 19.76 & 12.90          & 0.88 & 0.73                                       \\
           Abst.                   & \citet{zajic2004bbn}  \     & 25.12                         & 6.46                         & 20.12                                                          & 10.02   & $-$ & $-$        \\
           Abst.                    & \citet{baziotis2019seq}      & 22.13                         & 6.18                         & 19.30                                                        & 14.11   & 0.88 & 0.71           \\
           
           Ext.                    & \citet{west2019bottlesum}  \ & 22.85                         & 5.71                         & 19.87                                                         & 13.29    & $-$ & $-$          \\
           Ext.                    & \citet{schumann2020discrete}                            & 26.13                         & 7.98                         & 22.88                             & 4.73      & 0.86 & \textbf{0.94}                                \\
           Ext.                    & \citet{su2021non}                   & 26.26                         & 7.66                         & 22.83                                                  & 4.97     & 0.84 & 0.90               \\                  
           Ext.                  & \citet{liu-etal-2022-learning}              & 26.71                         & 7.68                        & 23.06                                             & 4.24   & 0.54 & 0.82                   \\
           Ext.                  & \citet{liu-etal-2022-learning}$^\dagger$              & 26.28                         & 8.11                        & 22.93                                             &    4.41  & 0.86 & 0.91                 \\

    Abst. & \cellcolor[HTML]{EFEFEF}$\text{\MD}$ \hspace{.01ex}   & \cellcolor[HTML]{EFEFEF}\textbf{27.88} & \cellcolor[HTML]{EFEFEF}\textbf{9.35} & \cellcolor[HTML]{EFEFEF}\textbf{24.49} & \cellcolor[HTML]{EFEFEF}0.0 & \cellcolor[HTML]{EFEFEF}\textbf{0.90} & \cellcolor[HTML]{EFEFEF}0.89  \\ 
    
    Abst.                    & \MD{} w/o RL                    &     24.66                     &      7.69                    &     21.90       &    7.48 & 0.88 & 0.79     \\

    \bottomrule
    
    \end{tabular}}
    \caption{Automatic evaluation on DUC2004 dataset. FD and FL stand for the fidelity and fluency, respectively.}
    \label{tab:duc}
\end{table}

\subsubsection{Metrics and Evaluation Protocol}
We use ROUGE, a word-overlapping ratio between generated and human-written summaries: ROUGE-\textit{n} for \textit{n}-gram matching and ROUGE-L for longest common subsequence matching. We use ROUGE F-1 (RF) on the Gigaword dataset, but use ROUGE recall (RR) on the DUC2004 dataset by following its evaluation protocol. 
In addition, we measure the \textit{fidelity} (i.e., content preservation) of generated summaries to input texts using SentenceBERT\footnote{We use cosine similarity between input texts and generated summaries, which are embedded by SentenceBERT.} \cite{reimers-gurevych-2019-sentence} and the \textit{fluency} of generated summaries with a language model, i.e., GPT-2 \cite{radford2019language}, based on Equation \ref{eq:R_F}.

Besides, since ROUGE gets higher as the summary gets longer, we group models based on the average length of the generated summaries for fair comparisons by following \citet{schumann2020discrete,liu-etal-2022-learning}.
We consider both settings of summarizing with a condition of a length (i.e., 8, 10, 13 words) and compression ratio (i.e., 50\% of the length of input texts) as done in the prior work. 
We also note that the evaluation protocol of DUC2004 truncates summaries that exceed 75 characters for fair comparisons in terms of the summary length. 

\subsubsection{Models Compared}

\textbf{Abstractive models }
\citet{zajic2004bbn} summarize texts using a syntax tree trimming. 
\citet{wang2018learning} train a model with an adversarial and cycle consistency loss. 
\citet{fevry2018unsupervised} utilize a denoising autoencoder.
\citet{zhou2019simple} model fidelity and fluency of summaries via contextual matching.
\citet{baziotis2019seq} stack autoencoders to impose the cycle consistency loss.

\noindent \textbf{Extractive models }
\text{Lead} baseline truncates texts from the beginning to the target lengths. 
\citet{west2019bottlesum} iteratively delete words from a text to generate a summary based on a fluency score. 
\citet{schumann2020discrete} search for the best word combination from texts based on a hill-climbing algorithm.
\citet{liu-etal-2022-learning} train a non-autoregressive transformer using summaries generated by \citet{schumann2020discrete} with corresponding input texts in a supervised manner. 
We also report another non-autoregressive model \cite{su2021non} that is trained similarly to \citet{liu-etal-2022-learning}.

\subsubsection{Implementation Details}
\label{sec:imp_detail}
We use sent2vec \cite{pgj2017unsup} as a word embedding-based projection function (i.e., $f$ in Equation \ref{eq:reward_cp}) that is trained  on the text corpus (i.e., news articles) by following the prior work \cite{schumann2020discrete,liu-etal-2022-learning}. We also report the results of \MD{} with SentenceBERT as a BERT-based projection function (Section \ref{sec:abl}).
As a language model, we use pretrained GPT-2 to obtain the fluency reward ($\psi$ in Equation \ref{eq:reward_fl}). 
We fine-tuned the language model on a target corpus (i.e., news headlines) as done in prior work \cite{zhou2019simple,schumann2020discrete}. 
As a policy $\boldsymbol{\pi}_\theta$, we use pretrained T5 \cite{raffel2020exploring}.
For the multi-summary learning mechanism, we train \MD{} with a set of lengths $L=\{8, 10, 13\}$ for the length-based evaluation and with a set of compression ratios $L=\{30\%, 40\%, 50\%\}$ for the compression ratio-based evaluation.
During beam search, we select a summary that maximizes the rewards (i.e., $\mathcal{R}_C, \mathcal{R}_F, \mathcal{R}_L$) and does not include predefined patterns. We provide more details in Appendix \ref{apnd:imp}. 

\begin{table}[]
    \centering
    \resizebox{0.8\linewidth}{!}{
    \begin{tabular}{l|ccc|c|ccc}
    \toprule
        & \multicolumn{4}{c|}{Majority} & \multicolumn{3}{c}{Unanimity} \\
        \multirow{-2}{*}{Criteria} & Win & Tie & Lose & $\kappa$ &  Win & Tie & Lose  \\ \midrule 
        \multicolumn{8}{c}{Comparison to \citet{schumann2020discrete}} \\ \midrule
        Fidelity  & \blue{52}  & 31  & \red{17} & 0.33   & \blue{26}  & 10  & \red{3}  \\
        Fluency & \blue{32} & 58  & \red{10} & 0.22   & \blue{13} & 9  & \red{5}  \\ \midrule
        \multicolumn{8}{c}{Comparison to \citet{liu-etal-2022-learning}} \\ \midrule
        Fidelity  & \blue{69}  & 4  & \red{27} & 0.59   & \blue{53}  & 3  & \red{16}  \\
        Fluency & \blue{69} & 10  & \red{21} & 0.51  & \blue{50} & 2  & \red{10}  \\
      
    \bottomrule
    \end{tabular}
    }
    \caption{Human evaluation results. $\kappa$ denotes Fleiss' kappa representing inter-annotator agreements.}
    \label{tab:human_eval}
\end{table}

\subsection{Automatic Evaluation}
We compare the summary quality of the models in Table \ref{tab:rouge} and \ref{tab:duc}, and make the following observations. \MD{} consistently shows the best \text{ROUGE} scores compared to both abstractive and extractive models over different groups of summary length. 
In terms of the \textit{fidelity}, \MD{} consistently achieves the best score compared to the baseline models, although \citet{schumann2020discrete,liu-etal-2022-learning} also consider the fidelity score ($\mathcal{R}_C$ in \MD{}) during training time.
\MD{} achieves competitive \textit{fluency} scores, while \MD{} is generally better than the best baseline model   \cite{liu-etal-2022-learning}. 

Moreover, as \citet{liu-etal-2022-learning} do not use a pretrained model, we include another baseline denoted by \citet{liu-etal-2022-learning}$^\dagger$ that uses the same initial model (i.e., pretrained T5) as \MD{} for fair comparisons. 
We observe that \MD{} still outperforms \citet{liu-etal-2022-learning}$^\dagger$ with the pretrained model. 

To investigate the effect of our RL framework, we consider \MD{} that is not trained under the RL framework (denoted by \MD{} w/o RL).  
The model is substantially inferior compared to \MD{} and the baseline models, indicating that our RL framework is vital to surpassing the recent extractive models. 

In a nutshell, \MD{} achieves the best ROUGE, fidelity, and competitive fluency thanks to our RL framework. 
We also observe that the inference time of \MD{} is competitively short compared to the state-of-the-art baseline models (Appendix \ref{apnd:inftime}).

\subsection{Human Evaluation}
We perform human evaluations to compare the summary quality between \MD{} and the baseline models, i.e., \citet{schumann2020discrete} and \citet{liu-etal-2022-learning}, on Gigaword data with 10 words as the summary length (Table \ref{tab:human_eval}). 
We provide the summaries generated by \MD{} and each baseline model along with the corresponding input texts to annotators, who are asked to choose a better summary in terms of fidelity and fluency. We ask a global annotation corporation to have three native speakers annotate 100 summaries.
We use majority voting and unanimity to consolidate the annotators' responses. We analyze the inter-annotator agreement based on Fleiss' kappa $\kappa$,\footnote{We follow \citet{landis1977measurement} to interpret kappa $\kappa$.} indicating \textit{fair} agreement for the comparison between \MD{} and \citet{schumann2020discrete} and \textit{moderate} agreement for the comparison between \MD{} and  \citet{liu-etal-2022-learning}.

In Table \ref{tab:human_eval}, \MD{} substantially outperforms the baseline models in both criteria. 
Particularly, the annotators indicate that \MD{} generates more fluent summaries than \citet{schumann2020discrete}, despite their highest fluency score in Table \ref{tab:rouge}. 
Such a discrepancy between automatic and human evaluation results has been also observed in recent work \cite{kuribayashi-etal-2021-lower}, and thus we argue that human evaluations are crucial for accurately evaluating the fluency of generated summaries.
From this experiment, we conclude that \MD{} is indeed superior to the baseline models based on both automatic and human evaluations.

\begin{table}[t]
\small
\centering
\renewcommand{\arraystretch}{1.1}
\scalebox{0.85}{
\begin{tabular}{c|cc|cc|cc}
\toprule
\multirow{2}{*}{Dataset}   & \multicolumn{2}{c|}{Gigaword} & \multicolumn{2}{c|}{Gigaword} & \multicolumn{2}{c}{DUC2004}  \\
                           & \multicolumn{2}{c|}{($l=8$)}  & \multicolumn{2}{c|}{($l=10$)} & \multicolumn{2}{c}{($l=13$)} \\ \midrule
Ratio                      & \multicolumn{2}{c|}{43.1\%}   & \multicolumn{2}{c|}{51.4\%}   & \multicolumn{2}{c}{60.4\%}   \\ \midrule
Avg. \# words              & \multicolumn{2}{c|}{1.29}     & \multicolumn{2}{c|}{1.35}     & \multicolumn{2}{c}{1.43}     \\ \midrule
\multirow{3}{*}{Top-3 POS} & IN           & 33\%         & IN          & 38\%         & IN           & 46\%        \\
                           & NNS          & 17\%         & NNS         & 17\%         & NNS          & 18\%        \\
                           & TO           & 14\%         & NN          & 12\%         & NN           & 10\%        \\
                        \bottomrule
\end{tabular}
}

\vspace{2px}

\scalebox{0.85}{
\begin{tabular}{c|l||c|l}
    \toprule
    Tag & Meaning & Tag & Meaning  \\ \midrule
    IN & Preposition or conjunction & NNS & Noun, plural  \\ 
    NN & Noun, singular or mass & TO & ``\textit{to}"   \\
    \bottomrule
\end{tabular}
}

\caption{Statistics of new words in summaries generated by \MD{} (top) and the meaning of POS tags (bottom). 
$l$ denotes the target summary length.
}
\label{tab:new_word}
\end{table}

\subsection{Frequency Analysis of New Words}
We demonstrate the benefit of \MD{} as an abstractive model in Table \ref{tab:new_word}.
In this experiment, we examine the generated summaries that contain new words, i.e., words that do not appear in input texts. We observe that the ratio of summaries that contain new words is around 50\%, and roughly 1.3 new words appear per summary. 
This result indicates that \MD{} frequently performs the abstractive operation (i.e., generating new words) so that \MD{} achieves higher summary quality than the extractive baselines, which merely select words from the input texts. 
We also report POS tags of new words, and observe that \MD{} mainly generates prepositions and nouns as new words.
We illustrate the generated summaries with new words in the following section \ref{sec:qa}.

\begin{table}[]
    \footnotesize
    \renewcommand{\arraystretch}{1.2}
    \centering
    \resizebox{0.999\linewidth}{!}{
    \begin{tabular}{c|cc|cc|cc}
    \toprule
       
        \multirow{2}{*}{Dataset}   & \multicolumn{2}{c|}{Gigaword} & \multicolumn{2}{c|}{Gigaword} & \multicolumn{2}{c}{DUC2004}  \\
                           & \multicolumn{2}{c|}{($l=8$)}  & \multicolumn{2}{c|}{($l=10$)} & \multicolumn{2}{c}{($l=13$)} \\ \midrule
        Metric & RF-1 & RF-L & RF-1 & RF-L & RR-1 & RR-L  \\ \midrule
        \MD{}&  	\ul{29.09}&	\ul{26.80}&	\ul{29.94}&	\ul{27.12}&	27.88&	\ul{24.49} \\ 
        $-$ MSL&  	28.49&	26.33&	29.79&	27.03&	27.57&	24.20 \\
        $-$ PTR& 	28.02&	25.91&	29.20&	26.55&	\ul{28.02}&	24.32 \\ \midrule 
        $-$ $\mathcal{R}_F$& 	27.28&	25.23&	27.99&	25.34&	26.30&	23.22 \\
        $-$ $\mathcal{R}_C$& 	26.31&	24.49&	27.82&	25.52&	25.97&	22.80 \\        
        $-$ $\mathcal{R}_C$ $+$ $\mathcal{R}_{AE}$& 	26.26 &	24.41 &	27.79 &	25.44 &	26.28 &	22.91 \\ 
        SBERT as $f$ &	\textbf{29.99}&	\textbf{27.56}&	\textbf{30.76}&	\textbf{27.93}&	\textbf{28.92}&	\textbf{25.25} \\

    \bottomrule
    \end{tabular}
    }
    \caption{Ablation study.}
    \label{tab:ablation}
\end{table}

\subsection{Ablation Study}
\label{sec:abl}
This section provides an ablation study to inspect the effect of each component in \MD{} (Table \ref{tab:ablation}).
We first train \MD{} without the multi-summary learning mechanism ($-$ MSL) and the prompt-based text reconstruction task ($-$ PTR), and observe that the performance generally degrades. Thus, both components are necessary to enhance the summary quality. In the following section \ref{subsec:msl} and \ref{subsec:ptr}, we provide in-depth analyses on each component.
We then train \MD{} without the fluency ($- \mathcal{R}_F$) and content preservation ($- \mathcal{R}_C$) reward, and observe that both rewards are essential to generating high-quality summaries. 

We further compare the semantic similarity and a word-level similarity adopted by prior abstractive models \cite{wang2018learning,baziotis2019seq}. 
By following their approaches, we build an autoencoder with an additional seq2seq model (i.e., pretrained T5). 
We then design a reward to minimize the reconstruction loss $\mathcal{L}_{AE}$ with a scaling factor, i.e.,  $\mathcal{R}_{AE} = \exp{(-\mathcal{L}_{AE}/\sigma_{AE})} \in (0,1]$.
\MD{} with the word-level reward, i.e., $-\mathcal{R}_C$ $+\mathcal{R}_{AE}$, substantially decreases ROUGE scores, indicating that the semantic similarity is more effective in capturing the core contents than the word-level similarity.
From the result, we show that the semantic similarity is a reason for the superior performance of \MD{} compared to prior abstractive models. 

Lastly, we replace the projection function $f$ from sent2vec with SentenceBERT (SBERT as $f$) and observe further improvements in ROUGE. This result implies that an accurate projection function $f$ can enhance the summary quality of \MD{}. 

\subsection{Effect of Multi-Summary Learning}
\label{subsec:msl}
We investigate which summary length benefits from the MSL mechanism in Figure \ref{fig:msl}. \MD{} tends to generate higher-quality summaries in the short length, i.e., 8 words, than our model not trained under the MSL mechanism (\MD{} w/o MSL). 
This result connotes that \MD{} can better learn to generate short summaries by referring to corresponding long summaries than independently generating short summaries. % during training time.  

\begin{figure}[t]
	\centering
	\includegraphics[width=.89\linewidth]{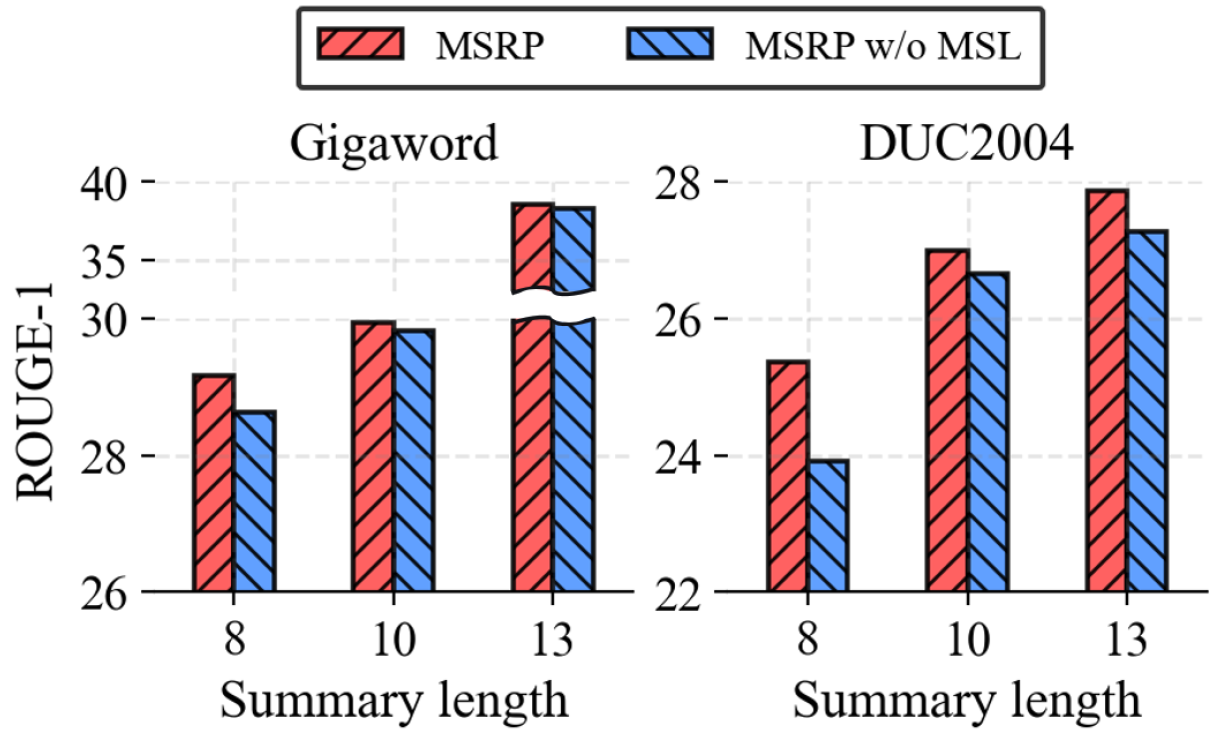} 
	\caption{Effect of multi-summary learning mechanism.}
	\label{fig:msl}
\end{figure}

\begin{figure}[t]
	\centering
	\includegraphics[width=.87\linewidth]{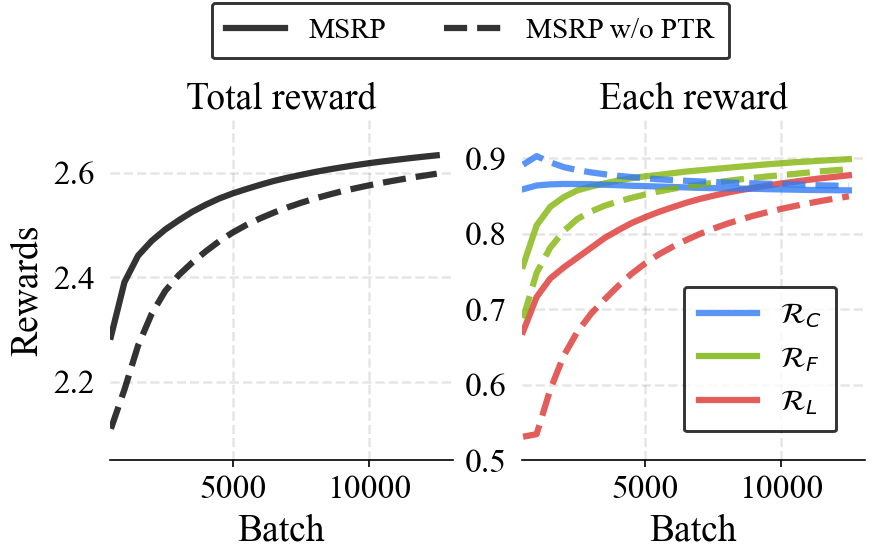} 
	\caption{Learning curve of rewards.}
	\label{fig:ptr}
\end{figure}

\subsection{Effect of Pretraining Task}
\label{subsec:ptr}
In Figure \ref{fig:ptr}, we inspect the effect of the PTR task. 
\MD{} more quickly optimizes the rewards (particularly the length reward $\mathcal{R}_L$) than the model not pretrained (\MD{} w/o PTR). 
This result implies that PTR task enables the model to learn how to control the summary length and summarize before RL training.
We thus posit that PTR task improves the summary quality as RL training takes advantage of the well-initialized model parameters.  

\begin{table}[]
    \fontsize{7.8}{8}\selectfont
    \centering
    \scalebox{0.96}{
        \begin{tabular}{p{1\columnwidth}}
        \toprule

        \textbf{\text{Input}}: israeli prime minister shimon peres said monday he was confident the ceasefire in lebanon would hold because it was in the best interests of both countries as well as syria .
          \\
                
        \vspace{0.05ex}
        
        \textbf{Reference}: peres confident ceasefire will hold \\        
                
        \midrule
        
        \textbf{\textsf{MSRP}}: israeli \blue{pm} confident ceasefire in lebanon \blue{will} hold \\
        
        \vspace{0.05ex}

        \textbf{NAUS}: israeli \red{minister} shimon peres confident ceasefire in Lebanon \\
        
        \vspace{0.05ex}
        
        \textbf{HC}: israeli \red{prime minister} shimon peres confident in syria \\
        \midrule \midrule

       \textbf{\text{Input}}:  president bill clinton announced reforms of the central intelligence agency aimed at restoring credibility in an espionage agency tarnished by the discovery of a russian mole in its midst .
          \\
        
        \vspace{0.05ex}
        
        \textbf{Reference}: clinton announces us intelligence reforms \\

        \midrule
        
        \textbf{\textsf{MSRP}}: president bill clinton \blue{announces} reforms of intelligence agency \\
        
        \vspace{0.05ex}

        \textbf{NAUS}: bill \red{reforms} intelligence agency aimed at restoring credibility  \\
        
        \vspace{0.05ex}
        
        \textbf{HC}: clinton \red{reforms} intelligence agency aimed at restoring credibility  \\
        
        \bottomrule
        
        \end{tabular}
    }
    \caption{Case study with generated summaries. NAUS: \citet{liu-etal-2022-learning}, HC: \citet{schumann2020discrete}.}
    \label{tab:examples}
\end{table}

\subsection{Case Study} 
\label{sec:qa}
We study the generated summaries to deeply understand the behavior and benefits of \MD{} compared to the best-performing baseline models (Table \ref{tab:examples}). 
In the top example, \MD{} generates an acronym \textit{pm} to replace \textit{prime minister}, a new word that does not appear in the input text. Similarly, \MD{} generates another new word, \textit{will}, resulting in a similar summary to the human-written summary that cannot be generated only by the extractive operation.

In the bottom example, \MD{} changes the past tense of the word \textit{announced} to the present tense \textit{announces}, which is more appropriate for news headlines than the past tense \cite{chovanec2003uses}. In contrast, the baseline models use \textit{reforms} as a verb, which can be reasonable. However, \MD{} preserves both important words \textit{announces} and \textit{reforms} so that the summary of \MD{} is more similar to the referential summary. We thus affirm that \MD{} surpasses the state-of-the-art extractive models by performing the abstractive operations for summarization. 

\section{Conclusion}
This work employs the RL for unsupervised abstractive sentence summarization with the rewards representing the summary quality and length. We invent the multi-summary learning mechanism to make the summaries with varying lengths mutually enhance each other. 
In addition, we design the prompt-based text reconstruction task to further improve the RL training.
Experimental results show that \MD{} achieves the state-of-the-art summary quality on both automatic and human evaluation.

\section*{Limitations}
RL enables summarization models to learn how to summarize with rewards representing the summary quality even though the rewards are non-differentiable. 
However, RL requires the word-sampling process to generate summaries in the training time. Thus, the computation time per input text is inherently longer than the sequence-to-sequence training with the cross-entropy loss, which the best baseline adopt \cite{liu-etal-2022-learning}. 

As a remedy, we expect non-autoregressive models can enhance the training efficiency of the RL framework by generating words in parallel instead of sequentially generating words, i.e., autoregressive generation. An issue of non-autoregressive models is the inferior quality of generated texts compared to the autoregressive models \cite{su2021non}, as non-autoregressive models are limited to consider the previously-generated words. Thus, future work can study  non-autoregressive models in the RL framework to enhance training efficiency while maintaining the summary quality.

It is worth noting that the total training time of \MD{} is shorter than the one of the best baseline \cite{liu-etal-2022-learning} despite the RL training. The best baseline depends on the summaries generated by \citet{schumann2020discrete}, while their inference time is excessively long due to the search operation. Based on the inference time in Appendix \ref{apnd:inftime}, 27 hours are required to generate summaries for 3M texts that are used by \citet{liu-etal-2022-learning}, while the training time of \MD{} with the pretraining task is about 8 hours. Thus, \MD{} is more efficient in terms of the total training time than the best baseline if we consider its data-generation time.

\section*{Acknowledgement}
We thank Dongha Lee, Seongbo Jang, and Seonghyeon Lee for their constructive comments.
This research was supported by IITP (No.2018-0-00584, SW starlab), (No.2019-0-01906, Artificial Intelligence Graduate School Program (POSTECH)), NRF grant (South Korea, No.2020R1A2B5B03097210, No.2021R1C1C1009081) funded by the Korea government (MSIT), the MSIT under the High-Potential Individuals Global Training Program (No.IITP-2020-0-01649), and it has been supported by Microsoft Research.

\bibliography{anthology,custom}
\bibliographystyle{acl_natbib}

\clearpage

\appendix

\section{Appendix}
\label{sec:appendix}
This appendix provides the details of our work.

\subsection{Shape of scaling factors}
\label{apnd:scale}
\begin{figure}[t]
	\centering
	\includegraphics[width=.99\linewidth]{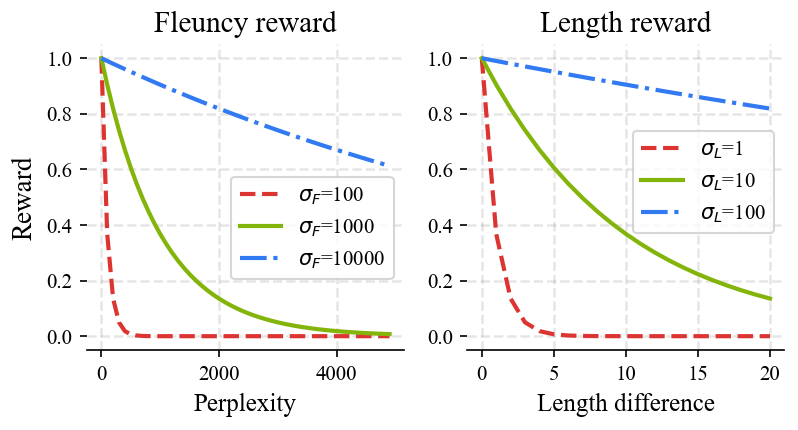} 
	\caption{Reward over different scaling factors.}
	\label{fig:scale}
\end{figure}

In Figure \ref{fig:scale}, we provide the shape of the fluency and length reward functions ($\mathcal{R}_F, \mathcal{R}_L$) over different values of the scaling factors. By tuning the scaling factors (Section \ref{apnd:tuning}), we observe that $\sigma_F=1000$ and  $\sigma_L=10$ produce the best result. This result indicates that smoother functions are desirable to train \MD{} than the original and steep functions, i.e., $\sigma_F=1, \sigma_L=1$. We note that the mean of the perplexity of GPT-2 on summaries is around 3,000. Low perplexity means high fluency. 
In addition, we do not apply the exponential function to the reward of content preservation ($\mathcal{R}_C$) as its range is bounded into $[0,1]$ by cosine similarity with normalization, i.e., $(\mathtt{cos}(\cdot, \cdot)+1)/2$.

\subsection{Details of Pretraining Task}
\label{apnd:ptr}
We provide the details of \textit{prompt-based text reconstruction} task (Table \ref{tab:pretrain}).
First, we shuffle a portion of words in a given text $\textbf{t}$ to make the model learn to \textit{reorder} the shuffled words into the original order. Second, we drop a small number of words to give the model the ability of \textit{adding} words. Lastly, we add words from another text $\textbf{t}'$ into the target text $\textbf{t}$, which enables the model to learn to shorten a given text by \textit{removing} words in it. The resulting text after the perturbations, i.e., $\tilde{\textbf{t}}$, and its clean text $\textbf{t}$ act as a text-summary pair.
We set the ratio of shuffling, dropping, and adding words to 10\%, 10\%, and 100\% of the number of words in input text $\textbf{t}$ after tuning them on the validation data.

To control output lengths, we specify the target length $|\textbf{t}|$ at the beginning of the perturbed text $\tilde{\textbf{t}}$:
\begin{align*}
    \tilde{\textbf{t}} = [p(|\textbf{t}|) ,\tilde{w}_1, \tilde{w}_2, \cdots, \tilde{w}_n]
\end{align*}
where $p(|\textbf{t}|)$ is the prompt in the form of `$|\textbf{t}|$:' (e.g., `$20$:' in Table \ref{tab:pretrain}), $\tilde{w}$ is a word in the perturbed text $\tilde{\textbf{t}}$, and $n$ is the length of the perturbed text $\tilde{\textbf{t}}$. 
Thus, by inserting the prompt to texts, the model can be explicitly informed about the target length for the original text.

\subsubsection{Comparison with BART}
In contrast to our task, BART \cite{lewis-etal-2020-bart} considers general language modeling, and thus it only covers the deletion operation among the three perturbations. 
We inspect the output of BART, and observe that BART shortens only 3\% of given texts, while the model pretrained on our proposed task can shorten all the given texts (100\%). What's even worse is that BART mostly generates the same texts with input texts (82\% of input texts). Thus, we claim that the proposed pretraining task is better at summarization than the pretraining task of BART.  
In addition, our pretraining task trains a model to control the output length based on the length prompt, whereas the pretraining task of BART does not include the length information. 

\subsection{Implementation Details}
\label{apnd:imp}
\subsubsection{Hyperparameters and Models}
\label{apnd:tuning}
We tune the hyperparameters of \MD{} based on RF-1 on the validation data such as the learning rate in $\{0.0001, \underline{0.00005}, 0.00001\}$\footnote{The best value for each hyperparamter is underlined.} with AdamW optimizer, the batch size in $\{16, \underline{24}, 30\}$, and the number of training data in $\{100\text{K}, \underline{500\text{K}}, 1\text{M}, 3.8\text{M}\}$. We set the weight for \textit{L}2 regularization to $0.01$. For rewards, we tune the scaling factors such as $\sigma_F$ in $\{100, \underline{1000}, 10000\}$ and $\sigma_L$ in $\{1, \underline{10}, 100\}$. 
We also tune $\lambda$ in $\{0.001, 0.005, \underline{0.01}, 0.05, 0.1\}$ and $\alpha$ in $\{0.0, \underline{0.3}, 0.5, 1\}$.

For transformer models, we use HuggingFace library \cite{wolf2020transformers}.  
As a language model, we use pretrained GPT-2 \cite{radford2019language}, which consists of 6 layers, to obtain the fluency reward ($\psi$ in Equation \ref{eq:reward_fl}).\footnote{Model ID in HuggingFace library: \texttt{distilgpt2}}
As a policy $\boldsymbol{\pi}_\theta$, we use pretrained T5 \cite{raffel2020exploring}, which consists of 6 layers for each encoder and decoder.\footnote{Model ID in HuggingFace library: \texttt{t5-small}} We select the small architectures to save GPU memory. We also use SentenceBERT in the public repository.\footnote{\scriptsize\url{https://github.com/UKPLab/sentence-transformers}}

\subsubsection{Beam Search}
\label{app:beam}
We perform beam search to generate summaries where the beam size is 20. We then select the best summary that maximizes the following scores:
\begin{align*}
    \small
    s(\textbf{y}) =  \mathcal{R}_C(\textbf{y}, \textbf{t}) + \mathcal{R}_F(\textbf{y}) + \mathcal{R}_C(|\textbf{y}|, l) %- \mathcal{P}(\textbf{y})
\end{align*}
where $s$ is the score for the generated summaries. The terms are the content preservation reward, the fluency reward, and the length reward with a target length $l$, respectively. Before computing the score $s$, we remove undesirable patterns from generated summaries. We use two types of patterns: 1) a preposition, interrogative pronoun, or conjunction such as \textit{to} or \textit{when} at the end of summaries, i.e., ungrammatical texts, and 2) a day of week such as \textit{monday}, i.e., less essential information. Refer to Table \ref{tab:pattern} for the patterns. In table \ref{tab:inftime}, we provide the effect of beam sizes, which implies \MD{} shows consistently higher ROUGE scores than the baseline models over different beam sizes, while \MD{} reaches the similar summary length to the that of baselines with around 20 beams. 

\begin{table}[]
    \fontsize{10.1}{8}\selectfont
    \centering
    \scalebox{0.8}{
    \begin{tabular}{l|p{0.8\columnwidth}}
    \toprule
        Patterns & Words \\
    \midrule
        Ending with &  in, at, to, on, the, 's, of, a, for, with, is, into, by, his, her, when, and, but \\ 
        \midrule
        Including & sunday, monday, tuesday, wednesday, thursday, friday, saturday  \\ 
     \bottomrule
    \end{tabular}
    }
    \caption{Undesirable patterns with the detailed words}
    \label{tab:pattern}
\end{table}

\subsection{Analysis on Inference Time}
\label{apnd:inftime}
Table \ref{tab:inftime} tabulates the inference time of \MD{} and the baseline models.
We note that the number of beams used by \citet{liu-etal-2022-learning} is 6.
\citet{schumann2020discrete} require the excessively-long generation time due to the exhaustive search in the inference time. \citet{liu-etal-2022-learning} reduce the generation time by training a model based on the outputs of \citet{schumann2020discrete} and using a non-autoregressive model. Similarly, we train \MD{} based on the rewards, and thus it generates summaries in short times while producing higher ROUGE scores than the baseline models. It is worth noting that the generation time of \MD{} is competitive to \citet{liu-etal-2022-learning} when we consider that 1) the model size of \MD{} is double of \citet{liu-etal-2022-learning} and 2) \MD{} is an autoregressive model while \citet{liu-etal-2022-learning} use a non-autoregressive model, which is faster than autoregressive models. 

\begin{table}[]
    \centering
    \resizebox{0.999\linewidth}{!}{
    \begin{tabular}{c|cccc|c}
    
    \toprule
    Model		 & RF-1	& RF-2 & RF-L &	Len.	    & Inf. Time  \\ \midrule
   \citet{schumann2020discrete} 	& 27.03	& 10.13	& 24.61 &	9.8	    & 33.214  \\
    \citet{liu-etal-2022-learning}	& 28.55	& 9.97	& 25.78 &	9.8	    & 0.043  \\ \midrule
    \MD{} ($|B|=1$) 		        & 29.63	& 11.83	& 26.88 &	11.0    & 0.004  \\
    \MD{} ($|B|=2$)	                & 30.29	& 12.28	& 27.56 &	10.2    & 0.018  \\
    \MD{} ($|B|=5$)	                & 30.08	& 12.08	& 27.35 &	10.1    & 0.031  \\
    \MD{} ($|B|=10$)	            & 30.03	& 12.00	& 27.25 &	10.0    & 0.060  \\
    \MD{} ($|B|=20$)	            & 29.94	& 11.86	& 27.12 &	9.9     & 0.095  \\
    \MD{} ($|B|=25$)	            & 29.80	& 11.83	& 26.99 &	9.8	    & 0.121  \\ 

    \bottomrule
         
    \end{tabular}}
    \caption{Evaluation with inference time per text in second on Gigaword with a length of 10. $|B|$: beam size.}
    \label{tab:inftime}
\end{table}

\subsection{Pseudocode of \MD{}}
Algorithm \ref{algo:rl} describes the pseudocode of \MD{} during the training time. 
We note that \MD{} generates summaries without referring to other summaries with varying lengths in the inference time. 

\subsection{Reproducibility}

\begin{table}[h]
    \small
    \centering
    \resizebox{0.8\linewidth}{!}{
    \begin{tabular}{c|l}
    \toprule
        Type & \MD{} trained with Sent2Vec as $f$ \\ \midrule        
        Length & \url{anonsubms/msrp_length} \\
        Ratio & \url{anonsubms/msrp_ratio} \\ \midrule \midrule        
         Type & \MD{} trained with SentenceBERT as $f$ \\ \midrule
        Length & \url{anonsubms/msrp_length_sb} \\
        Ratio & \url{anonsubms/msrp_ratio_sb} \\
    \bottomrule
    \end{tabular}}
    \caption{Model ID of \MD{} in HuggingFace library.}
    \label{tab:link}
\end{table}

We provide our source codes and data in \textbf{\textit{msrp.zip}} for reproducing the experimental results. We also upload our models to HuggingFace library, enabling anyone to use \MD{} with a few lines of code. Refer to the uploaded model in Table \ref{tab:link}. 
The type \textit{Length} indicates \MD{} trained for length-based evaluation (Group A, B, and D in Table \ref{tab:rouge} and \ref{tab:duc}) and type \textit{Ratio} is \MD{} trained for compression ratio-based evaluation (Group C in Table \ref{tab:rouge}).

\begin{algorithm}[]
		\footnotesize
		\SetAlgoLined
		\SetKwInOut{Input}{Input}
		\SetKwInOut{Output}{Output}
		\Input{A policy $\boldsymbol{\pi}_\theta$, a set of texts $T$, a set of lengths $L$, a learning rate $\eta$, a training type $Type$}
		\Output{A trained policy $\boldsymbol{\pi}_\theta$}
		\DontPrintSemicolon
		\While{Convergence}{
		    \ForEach{$\textbf{t} \in T$}{
		        \If{$Type=$ multi-summary learning}{
		            $Y = \{\emptyset \}$ \\
    		        \ForEach{$l\in L$}{
    		            $\textbf{y}^l\: \small{\sim}\: \boldsymbol{\pi}_\theta(\cdot|l, \textbf{t})$ \\
    		            $Y = Y \cup \{\textbf{y}^l\}$
    		        }
    		         $\mathcal{J}(\theta)=- \mathbb{E}_{Y \small{\sim} \boldsymbol{\pi}_\theta(\cdot|L, \textbf{t})} \mathcal{R}^\star(Y, \textbf{t}).$ \\
		        }
		      
		        \Else{	               
		            $l\: \small{\sim}\: L$  \Comment*[r]{Sample a length}
    		        
		            $\textbf{y}^l\: \small{\sim}\: \boldsymbol{\pi}_\theta(\cdot|l, \textbf{t})$ 
    		        
    		         $\mathcal{J}(\theta)=- \mathbb{E}_{\textbf{y}^l \small{\sim} \boldsymbol{\pi}_\theta(\cdot|l, \textbf{t})} \mathcal{R}(\textbf{y}^l, \textbf{t}, l).$ \\
		        }
		        $\theta \leftarrow \mathtt{optimizer}(\theta, \bigtriangledown_\theta \mathcal{J}(\theta), \eta) $
		    }
		}
		\caption{Pseudocode of \MD{}}
		\label{algo:rl}
\end{algorithm}

\end{document}